\documentclass{article}

\usepackage{microtype}
\usepackage{graphicx}
\usepackage{subfigure}
\usepackage{booktabs} 
\usepackage{color}
\usepackage{graphicx}
\usepackage{multirow}
\usepackage{multicol}
\usepackage{amsmath}
\usepackage{hyperref}
\usepackage{comment}
\usepackage{stfloats}

\usepackage[accepted]{sysml2019}

\sysmltitlerunning{CaTDet: Cascaded Tracked Detector for Efficient Object Detection from Video}

\begin{document}

\twocolumn[
\sysmltitle{CaTDet: Cascaded Tracked Detector for Efficient Object Detection from Video}



\sysmlsetsymbol{equal}{*}

\begin{sysmlauthorlist}

\sysmlauthor{Huizi Mao}{stanford}
\sysmlauthor{Taeyoung Kong}{stanford}
\sysmlauthor{William J. Dally}{stanford,nvidia}

\end{sysmlauthorlist}

\sysmlaffiliation{stanford}{Stanford University}
\sysmlaffiliation{nvidia}{Nvidia Corporation}

\sysmlcorrespondingauthor{Huizi Mao}{huizimao@stanford.edu}
\sysmlcorrespondingauthor{William J. Dally}{dally@stanford.edu}

\sysmlkeywords{Vision, Video, Efficient Algorithms}

\vskip 0.3in

\begin{abstract}
Detecting objects in a video is a compute-intensive task. In this paper we propose CaTDet, a system to speedup object detection by leveraging the temporal correlation in video. CaTDet consists of two DNN models that form a cascaded detector, and an additional tracker to predict regions of interests based on historic detections. We also propose a new metric, mean Delay(mD), which is designed for delay-critical video applications. Experiments on the KITTI dataset show that CaTDet reduces operation count by 5.1-8.7x with the same mean Average Precision(mAP) as the single-model Faster R-CNN detector and incurs additional delay of 0.3 frame. On CityPersons dataset, CaTDet achieves 13.0x reduction in operations with 0.8\% mAP loss. 
\end{abstract}
]



\printAffiliationsAndNotice{}  

\section{Introduction}

We consider the task of detecting objects in a video. Video is an important data source for real-world vision tasks such as surveillance analysis and autonomous driving. Such tasks require detecting objects in an accurate and efficient manner.

Processing video data is compute-intensive. A \$50 camera can generate 1080p video stream at 25fps, while a \$1000 Maxwell Titan X with SSD512 algorithm can only detect objects at 19 fps~\cite{liu2016ssd}.

One approach to reducing the computational workload of video processing is to exploit the temporal and spatial locality of video: treating it as a sequence rather than running detection on each image separately. For most videos, the same object appears in adjacent frames(temporal locality) and in the nearby locations(spatial locality). 
We can therefore exploit this property to make object detection more efficient.

\begin{figure*}[!h]
  \centering
  \includegraphics[width=0.8\textwidth]{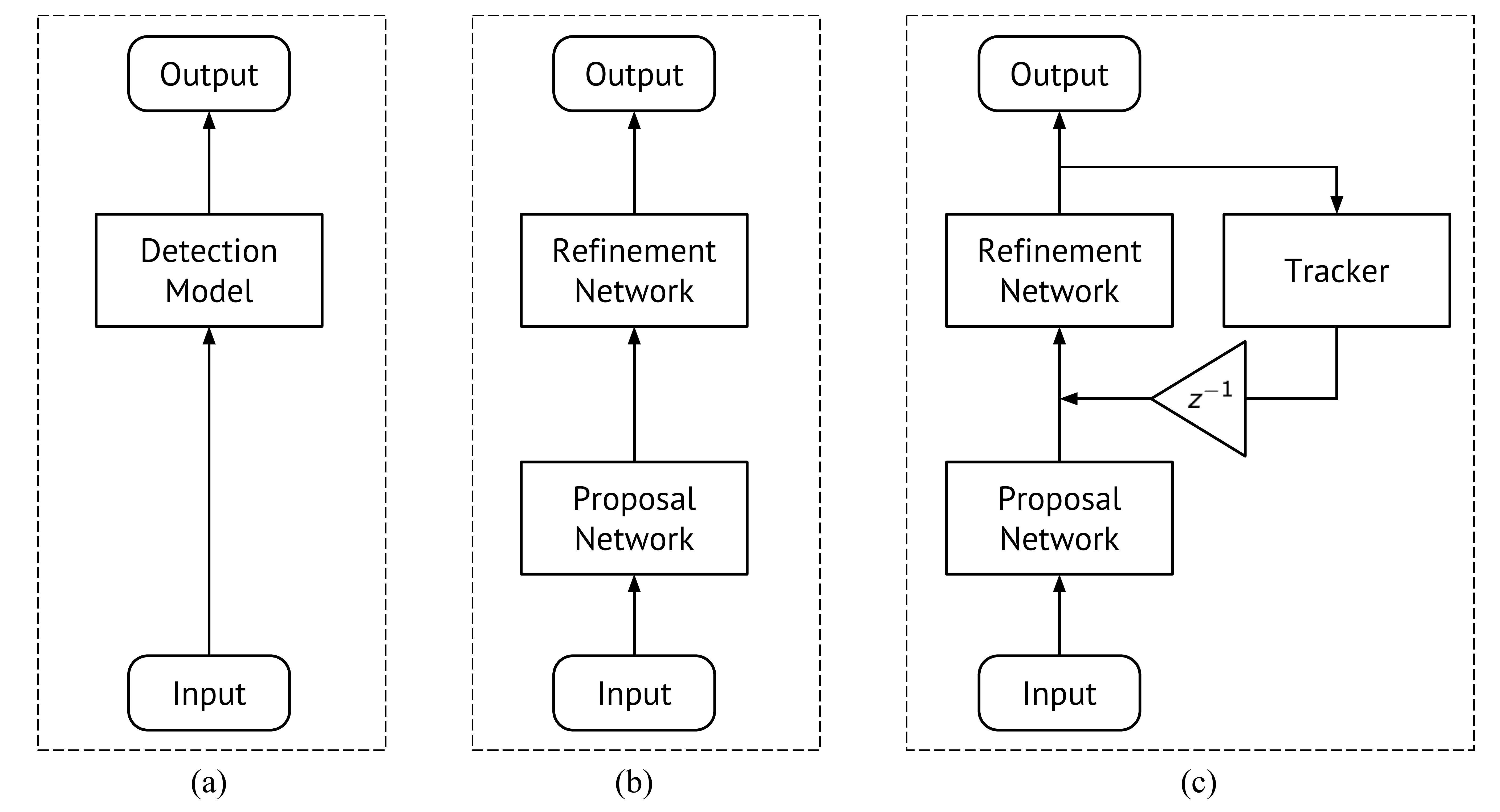}
  \caption{(a) Single-model detector system: the detection model is a CNN model for object detection, such as Faster R-CNN, SSD, etc. (b) Cascaded detector system: The cheap proposal network scans the whole images and produces potential object candidates to be calibrated by the Refinement Network. Both the proposal network and the refinement network are detection models, although technically the proposal network does not need to be. (c) CaTDet: a tracker is further added that incorporates previous frames' information to aid detection and save computations.}
  \label{fig:system}
\end{figure*}

We propose CaTDet (Cascaded Tracking Detector), a computation-saving framework for video detection that incorporates the tracker into a cascaded system. It is designed for but not limited to moving-camera delay-sensitive scenarios, e.g., autonomous driving. As shown in Figure~\ref{fig:system}, CaTDet is a detector cascade with temporal feedback. The tracker and the inexpensive proposal network extract the interesting regions in an image, which reduces the workload on the expensive refinement network.

For single-image object detection algorithms, various metrics such as Average Precision~\cite{everingham2010pascal} have been proposed to measure the detection quality. These metrics do not account for the temporal characteristics of a video. 
For many real-world video applications such as autonomous driving, the metric that matters is {\em delay}, the time from when an object first appears in a video to when it is detected.

Our contribution in this work is two-fold: the delay metric for object detection in video and CaTDet, a detection system to efficiently detect objects with the aid of temporal information. We evaluate CaTDet on KITTI~\cite{kitti} and CityPersons~\cite{citypersons} datasets, with both the traditional mAP metric and the new delay metric. The results show that CaTDet is able to achieve 5.1-8.7x speed-up with no loss of mAP and only a small increase in delay.

\section{Related Work}

\textbf{Object detection from video.}
Others have also exploited the temporal properties of video to improve object detection.
T-CNN~\cite{kang2016t} regularizes detection results with tracking output. 
In Detect and Track~\cite{dt}, an end-to-end architecture is proposed which incorporates detection and tracking to improve accuracy. Both methods require future frames to predict current frame, therefore they are non-causal.
Deep feature flow~\cite{dff} exploits temporal redundancy between neighboring frames by estimating optical flow with FlowNet~\cite{ilg2017flownet}.
It achieves high speed-ups by skipping feature extraction for non-key frames. 
Flow-guided feature aggregation~\cite{fgfa}, on the other hand, tries to improve accuracy by aggregating features from nearby frames.
In ~\cite{thp},  a unified approach of exploiting feature flow is presented, which improves both the detection accuracy and the speed. A similar flow based hardware solution is proposed in $ EVA^ 2$~\cite{buckler2018eva}.


\textbf{Region extraction and selected execution.} 
For videos with static backgrounds, only the changing regions require updates. CBinfer~\cite{cavigelli2017cbinfer} filters out moving pixels by change-based thresholding. In ~\cite{zhang2017kill}, foreground extraction methods are employed to extract regions of interest.  However, they do not work in the general case of a moving camera.

\textbf{Adaptive/cascaded neural network.}
Others also use a system-level approach to improve computation efficiency of Deep Neural Network. 
~\cite{chen2018optimizing} proposes a way to combine sparse expensive detection and dense cheap detection via a scale-time lattice. 
In Adaptive Neural Networks~\cite{adaptive}, a system is proposed that adaptively chooses the parts of a neural network model according to the input. 
SACT~\cite{sact} further develops this work~\cite{adaptive} by adaptively extracting high-probability regions and selectively feeding these regions into a deeper network.
While these works improve computational efficiency, they are designed for still image object detection which does not take advantage of temporal information.
NoScope~\cite{noscope} proposes a framework with selective processing and cascaded models that can speedup frame classification in video up to 1000x. However, the task of NoScope is relatively simple and the input is limited to static background videos.

\section{Proposed Detection System}

Figure~\ref{fig:system} illustrates a single-model detection system, a cascaded system, and CaTDet. The single-model system is a CNN model such as Faster R-CNN~\cite{faster} or SSD~\cite{ssd}. The cascaded model consists of two detection DNNs, a lightweight \textit{proposal network} followed by a higher-precision \textit{refinement network}. The proposal network selects regions of interest for the refinement network to reduce computation. Our proposed system, CaTDet, improves on the cascaded system by using a tracker to further extract regions of interest based on historical objects. 

CaTDet works in the following steps. The proposal network, which may be small and inaccurate, inputs every full video frame and outputs the potential locations of objects within the frame. The tracker tracks the history of high-confidence objects in the current frame and predicts their location in the next frame. For each frame, the outputs of the proposal network and the tracker are combined before being fed into the refinement network, and thereby we obtain the calibrated object information. 

\begin{figure*}[]
  \centering
  \includegraphics[width=\textwidth]{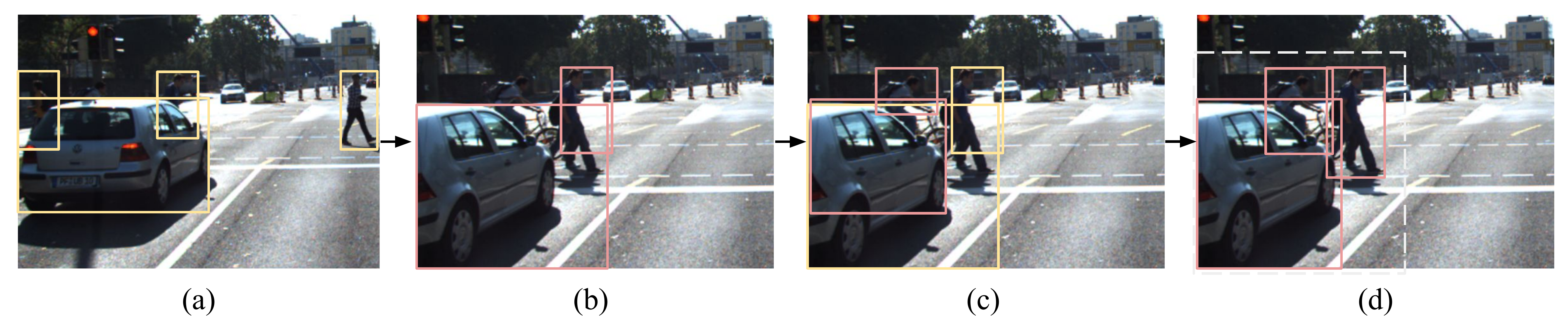}
  \vspace{-5pt}
  \caption{A clip of CaTDet's execution loop. Each step's new output is in red, while the output of previous steps is in yellow. (a) Last frame’s detections. (b) Locations of the pre-existing objects are predicted by the tracker. Objects leaving the current frame are deleted. (c) New detections from the proposal network are added. (d) The refinement network performs selected calibration, as shown in the dotted box. Duplicated detections will be deleted in the NMS step. }
  \label{fig:example}
\end{figure*}

CaTDet is based on the concept that validation and calibration are easier than re-detection. 
An object that has been detected in one video frame is very likely to appear in the next frame at a nearby location.  
Based on this prior, running a high-precision detector on regions around previous objects saves workload while preserving  accuracy.  This simple concept, however, faces two problems in practice:


\textbf{New objects}. It is difficult to predict where new objects will appear in a new video frame. To detect them requires scanning the entire image. We use a fast, inaccurate detector, the proposal network, to locate potential objects. A high false positive rate of the proposal network may be tolerated, as we have a more accurate detector to refine the results. By functionality, the proposal network is very similar to the region proposal network(RPN) in Faster R-CNN. 

\textbf{Motion and occlusion}. The movements of objects or the camera cause the locations to drift. In some cases due to occlusion, objects may be temporally invisible. If the refinement network simply looks at previous locations, minor mismatches or temporal misses could lead to permanent loss. Therefore, we use a tracker as a robust future location predictor.
For the tracker, standard tracking algorithms can be directly applied, except that the output is predicted locations instead of tracklets.

An example of CaTDet's workflow is given in Figure~\ref{fig:example}. Starting with the last frame's detection results in stage \textit{a}, the tracker updates its internal state and predicts the corresponding locations/proposals for the next frame as in stage \textit{b}. Together with the proposal net's outputs in stage \textit{c}, the proposals are fed into the refinement network to be calibrated. As shown in stage \textit{d}, the refinement network operates only in the regions of interest, thus saves the overall workload.

\section{Implementation Details}
In this section we present a detailed description of each module of CaTDet.

\subsection{Tracker}
In CaTDet, the tracker matches the objects in consecutive frames, estimates motion information and predicts next-frame locations. Notice that, the goal of typical object tracking problem is different, though the algorithm could be almost the same.
Tracking algorithms usually output tracked sequences of detected objects, and the predicted locations are intermediate results. The predictions, which indicate regions of interest, are then fed into the refinement network.

\begin{figure*}[!h]
  \centering
   \includegraphics[width=0.65\textwidth]{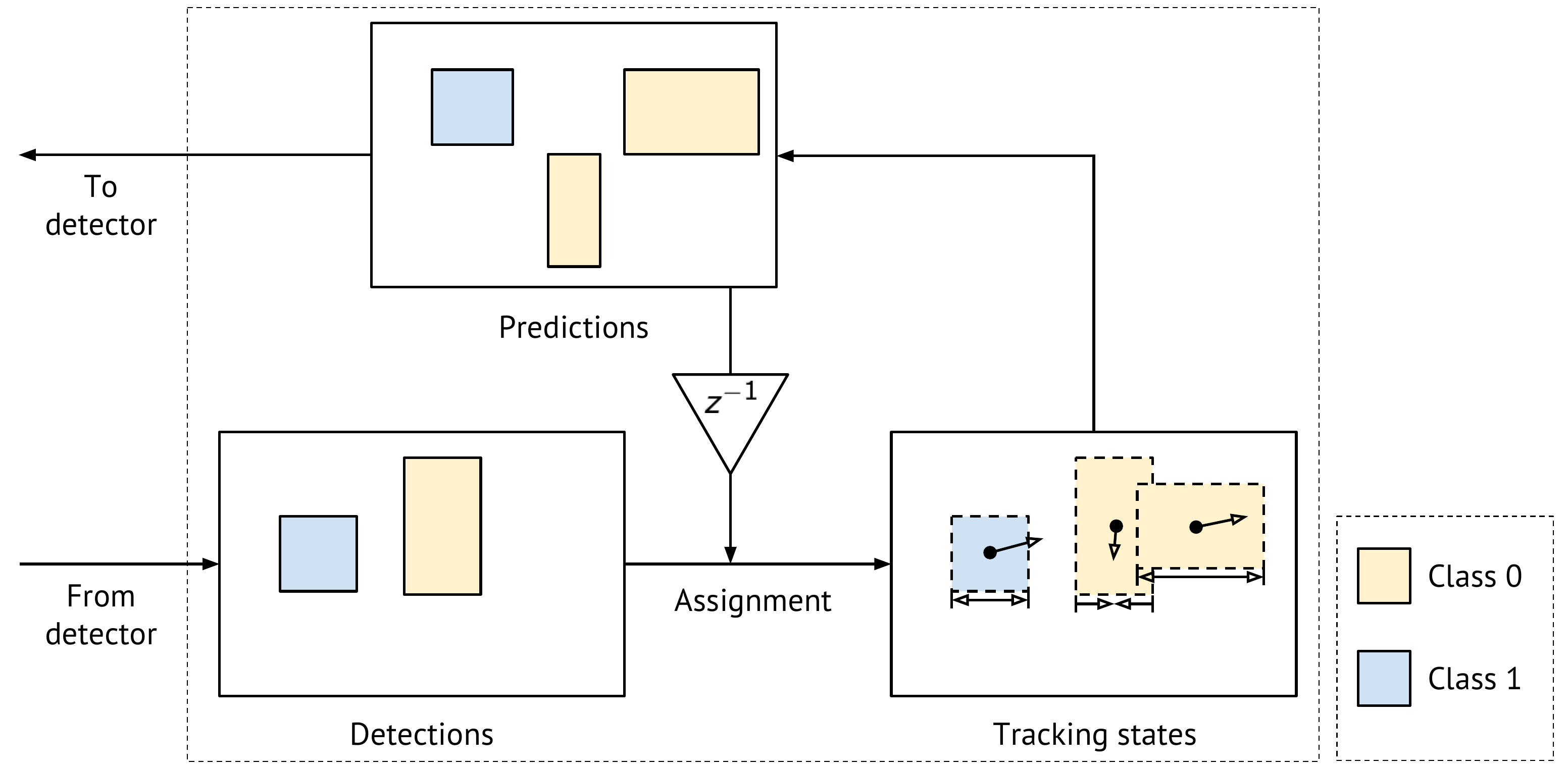}
  \caption{Tracker work flow. The tracker takes as input the current frame's detected objects (Detections) and outputs the predicted locations (Predictions) in the next frame. The predicted objects and detected objects of the current frame are matched to update the internal state (Tracking states), which includes the coordinates, sizes and motions of objects.}
  \label{fig:tracker}
\end{figure*}

Our tracking algorithm is inspired by SORT(Simple Online and Realtime Tracking)~\cite{bewley2016simple}. In SORT, two major components are object association and motion prediction. Object association matches the objects in two(or more) adjacent frames. Motion prediction uses the states of existing tracked objects to predict their locations in future frames. These two steps are executed iteratively for every new frame.

\textbf{Object association} module employs the modified Hungarian algorithm, just as SORT does, to match objects in two adjacent frames. For an N-to-M matching problem, we construct an N-by-M cost matrix of which the elements are negative Intersect over Union(IoU). For IoUs that are smaller than a threshold $\beta$, the two bounding boxes are set as non-relevant regardless of Hungarian algorithm's results. The output of object association is matched objects, lost objects (unmatched objects in the previous frame) and emerging objects (unmatched objects in the new frame). Notice that this process is performed one time per class.

\textbf{Motion prediction} module employs the simple exponential decay model instead of the Kalman Filter algorithm in SORT. It is observed to be more robust to different settings of frame rate and image resolution. For Kalman Filter, the parameters need to be carefully tuned on the training dataset~\cite{bewley2016simple}, while in the decay model they are set as constants. 

In the tracker, the state of an object is represented by two vectors $\mathbf{x} = [x,y,s]$, $\mathbf{\dot x} = \dot x, \dot y, \dot s]$ and a scalar $r$. Here $x$ and $y$ are the center coordinates, $s$ is the width of the bounding box and $r$ is the aspect ratio(height to width). The following update rule is adopted in the tracker:

\begin{equation}
\mathbf{\dot x}_{n+1} = \eta \mathbf{\dot x}_{n} + (1-\eta)( \mathbf{x}_{n+1} -\mathbf{x}_{n})
\end{equation}
\begin{equation}
\mathbf{x'}_{n+1} = \mathbf{x}_{n} + \mathbf{\dot x}_{n}
\end{equation}
\begin{equation}
r'_{n+1} = r_n
\end{equation}

Here $\mathbf{x'}$ and $r'$ are the predicted values, which are later converted into bounding box coordinates and fed into the refinement network. If the object is not matched in the new frame (missing object), the motion is kept constant for several frames until it is discarded. For emerging objects, the motion vector is initialized as $\mathbf{0}$. 

In our implementation, the IoU threshold $\beta$ is set to 0 and decay coefficient $\eta$ to 0.7, although the tracker is robust to a wide range of $\eta$. 

The tracker itself operates very efficiently. Our experiments on KITTI dataset show that it is able to reach 1082fps with single-thread Intel E5-2620 v4. However, the number of predicted objects from tracker affects the workload of the refinement network. We filter out the objects that are too small(width smaller than 10 pixels) or have been largely chopped by the boundary. Instead of tracking the missing objects for a fixed number of frames, an adaptive scheme is adopted that every match adds to confidence with a upper limit and every miss reduces confidence. Once the confidence value goes below zero, the object is discarded.
Compared with the original tracking algorithm, our tracker is optimized to reduce the number of predictions to save total operations of the system.

\subsection{Proposal network}
The proposal network and the refinement network are object detectors that use similar algorithms but 
have different sizes and accuracy. 
The proposal network is much smaller than the refinement network.
Together form a cascaded detector to save operations. 
For simplicity, we use Faster R-CNN~\cite{faster} for both the proposal network and the refinement network in CaTDet. 
Faster R-CNN was proposed in 2015, yet still serves as a good baseline detection framework.

For the proposal network, the standard Faster R-CNN settings are adopted. The image is first processed by a Feature Extractor that shares the same convolutional layers as the classifier model. 
A Region Proposal Network (RPN) predicts 3 types of anchors with 4 different scales for each location. 
After Non-Maximum Suppression (NMS), 300 proposals are selected and fed into the classifier. 
Our proposal network produces class labels, but they are unused.  

Four different types of simple ResNet models are used in our experiments, including the standard ResNet-18, which is our largest model. 

\begin{table}[h!]
\centering
\caption{Model specifications for proposal nets. In ResNet-18, all blocks are repeated 2 times. The number of arithmetic operations is measured with Faster R-CNN on KITTI dataset, of which the input image size is 1242x375 and the number of proposals is 300.}
\label{tab:models}
\vspace{2pt}
\begin{tabular}{l|llll}
  & ResNet- & ResNet- & ResNet- & ResNet- \\
  & 18& 10a& 10b& 10c\\
\hline
conv1      & 64        & 48         & 32         & 24         \\
block1     & 64(x2)    & 48         & 32         & 24         \\
block2     & 128(x2)   & 96         & 64         & 48         \\
block3     & 256(x2)   & 168        & 128        & 96         \\
block4     & 512(x2)   & 512        & 256        & 192       \\
\hline
ops & 138.3G & 20.7G & 7.5G & 4.5G \\
\hline
\end{tabular}
\vspace{-10pt}
\end{table}

\begin{figure*}[!h]
  \centering
  \includegraphics[width=0.8\textwidth]{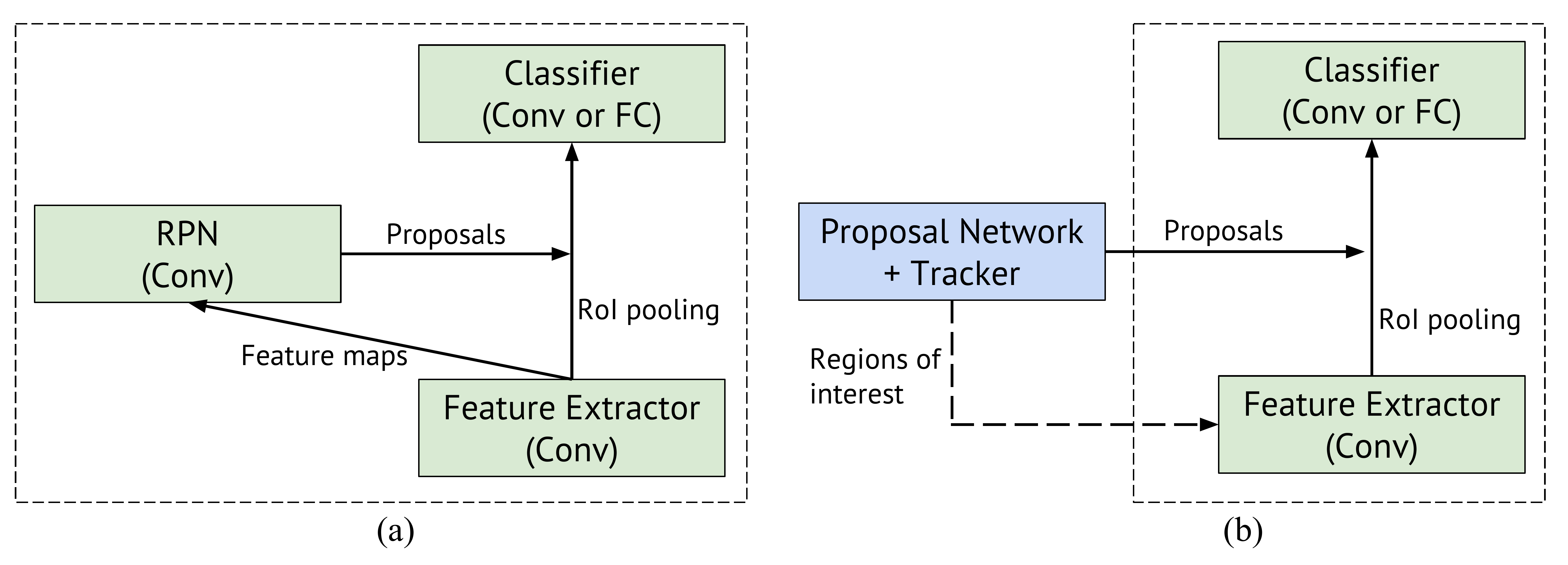}
  \caption{Compare the inference-time work flows of the standard Faster R-CNN model and the refinement network. (a) Standard Faster R-CNN detector: the RPN takes the feature maps from the feature extractor and . (b) Faster R-CNN detector for the refinement network: the proposals from the proposal network and the tracker instruct the feature extractor to only compute features on regions of interest. Regions of interest is a mask of all proposals over the frame. }
  \label{fig:refinement}
\end{figure*}

\subsection{Refinement network}
The refinement network runs a modified version of Faster R-CNN. Two major differences are listed as follows:

\textbf{Selected regions of features}. The proposal network and the tracker together provide a high-recall region selection, so the refinement network only requires parts of the feature maps that are corresponding to the selected regions. A margin of 30 pixels is appended around the proposals to maintain enough information for the ConvNet. Here we are interested in the real number of operations needed to extract required features, which is platform-independent, therefore the regions-of-interest are not required to be rectangular. Some works~\cite{zhang2017kill} merge regions into rectangles to maximize the computational efficiency on GPU. 

\textbf{Reduced number of proposals}. The typical number of proposals from RPN is 300~\cite{faster}. In CaTDet, however, it is usually much smaller, as we observed that the proposal network and tracker provide much more accurate proposals than RPN. That helps save computations for models like ResNet-50, of which the classifier is a relatively heavy ConvNet. 

Apart from its own hyper-parameters, the workload of the refinement net is also affected by the hyper-parameters of the tracker and the proposal net. There are two that significantly affect the inference speed -- the confidence thresholds for the tracker's input and the proposal network's output. Due to the fact that the proposal net and the tracker predict overlapping proposals, their impact on total system operations is coupled. Increasing either threshold will decrease the total number of operations, while run the risk of hurting accuracy.

\section{Evaluation Metrics} 
We selected two metrics to evaluate our detection results, mean Average Precision (mAP) and mean Delay(mD). 

\begin{figure}[h]
  \centering
  \includegraphics[width=0.48\textwidth]{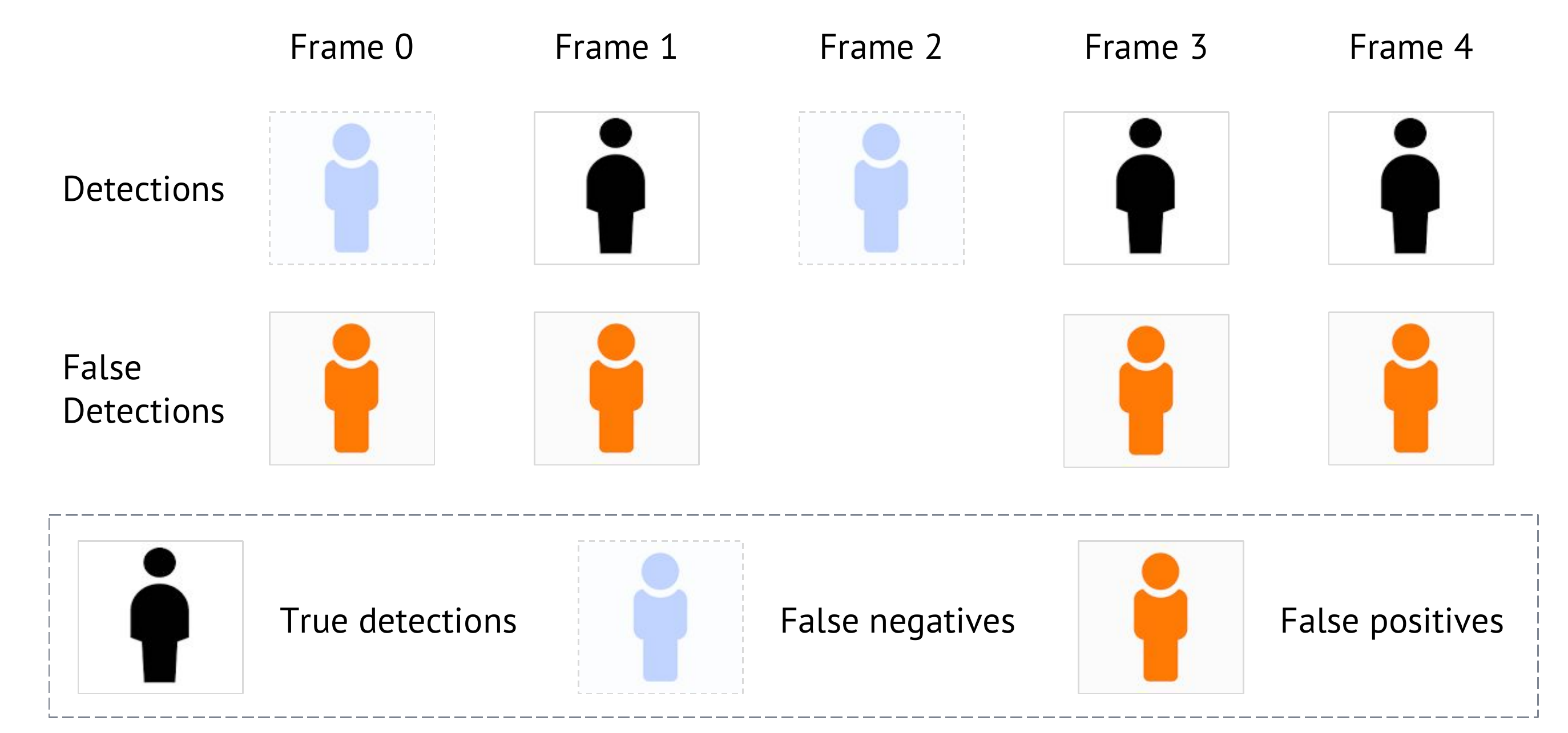}
  \vspace{-10pt}
  \caption{Illustration of how recall, precision and delay correspond to ground truth and to detections. One ground-truth object spans 5 frames in this example. There are in total 7 detections (3 true detections and 4 false positives) and 2 false negatives. Only the single false negative in Frame 0 counts towards delay. Recall = 3/5. Precision = 3/7. Delay = 1. }
  \label{fig:delay_illustration}
\end{figure}

\textbf{Average Precision (AP)} is a common metric to measure the overall quality of detections. It is defined as the integral of precision over recall, while in practice, precisions at discrete recall values are selected to approximate the integral. As an example, in Pascal VOC~\cite{everingham2010pascal}, 11 recall values ranging from 0 to 1.0 are averaged to calculate AP. Mean Average Precision is simply the arithmetic mean of all different classes' AP values. It is comprehensive enough to capture a detector's accuracy regarding the precision-recall trade-off, which makes it the standard evaluation metric in most detection benchmarks~\cite{everingham2010pascal, Imagenet, mscoco}.

mAP is designed in such a way that all detections are weighed equally, which is the typical case for single image detection. However, we argue that not all detections are equal in the video detection problem. Given detections of a single object in a video sequence, the first detection is particularly important as it determines the response time. For delay-critical scenarios like autonomous driving, it is essential to have a low response time. For applications that are not sensitive to detection delay, in-sequence detection errors can be fixed with post-processing~\cite{kang2016t}, while detection errors in the beginning are much more difficult to fix.  

\textbf{Mean Delay} is proposed to emphasize early detection of an instance. By definition, delay simply means the number of frames from the beginning frame of an groundtruth object sequence to the first frame the object is detected. An example is given in Figure~\ref{fig:delay_illustration}, where the relationship between recall, precision and delay is illustrated. By averaging the delay over all classes, we obtain mean Delay(mD). Notice that for simplicity, the delay metric does not take into consideration the frame rate of the video steam.

It is easy to define and measure the delay, however, it is difficult to compare the average delay across different methods. The reason is that one can always reduce the delay by detecting as many objects as possible. The AP metric makes a trade-off between false negatives (recall) and false positives (precision), while the delay metric only penalizes false negatives as shown in Figure~\ref{fig:delay_illustration}.

To fairly compare the average delay of different methods, we measure the delay at the same precision level. To be specific, given a target precision value $\beta$, we select the confidence threshold so that the mean precision of all classes matches $\beta$, thereby measure the average delay for different methods.

\begin{equation}
mD@\beta  := \frac{1}{|C|} \sum_{c \in C} Delay_c(t_\beta)\text{ , }
\end{equation}
$ \text{where } C \text{ is the set of classes and } t_\beta \text{ is chosen as}$
\begin{equation}
\frac{1}{|C|} \sum_{c \in C} Prec_c(t_\beta) = \beta
\end{equation}

The above formula only describes the entry delay. Some applications may also be sensitive to exit delay, which is defined as the actual exit frame minus the predicted exit frame. Due to the fact that entry frame is harder to predict as no prior knowledge exists, we are focusing on entry delay throughout this paper.

\section{Experiments on KITTI}

\subsection{Dataset overview} \label{kittidata}
KITTI is a comprehensive dataset with multiple computer vision benchmarks related with the autonomous driving task. Out of the available benchmarks, we use the data from the 2d object (detection) benchmark for training and and the tracking benchmark for evaluation

KITTI's 2d object benchmark contains 7481 training images. The tracking benchmark contains 21 training sequences with a total number of 8008 frames at a frame rate of 10 fps. 
Overlapping images exist in the 2d object dataset and tracking dataset, therefore we filtered out the duplicated images in the training set. After that there are 4383 remained images.
Both the detection and tracking datasets contain a relatively small number of images, compared with ImageNet DET (128k images) and MS-COCO (83k images).

Our results on KITTI dataset is evaluated using the official evaluation codes as described in the KITTI paper~\cite{kitti}.
Only  \textit{Car} and \textit{Pedestrian} classes are evaluated in accordance to the tracking subset.
Following KITTI's convention, an overlap of $50\%$ is required for a valid detection of \textit{Pedestrian}, while the class \textit{Car} requires at least $70\%$.
The open-sourced KITTI development kit can be found via the following link\footnote{\href{https://s3.eu-central-1.amazonaws.com/avg-kitti/devkit\_object.zip}{https://s3.eu-central-1.amazonaws.com/avg-kitti/devkit\_object.zip}}.

It should also be noticed that, in the official KITTI evaluation protocol, there are three difficulty levels (\textit{Easy}, \textit{Moderate}, \textit{Hard}). Each difficulty level sets specific thresholds of bounding box size, occlusion level and truncation for a valid ground truth. For example, in \textit{Easy} mode only ground truth objects that are wider than 40 pixels and fully visible are evaluated, otherwise they do not count towards false negatives. In our experiments, we find that the \textit{Easy} mode does not distinguish different methods, therefore we do not show its results in the following sections.

\subsection{Training procedure}
\label{sec:training}
Due to the fact that the training set is relatively small compared with the common datasets used for Deep Neural Networks training, we adopt a 3-stage training pipeline for the Faster R-CNN models. Similar to other detection frameworks, our models are first pretrained on the ImageNet classification dataset. The classification model (all layers minus the last classifier layer), together with the RPN module, is then trained on MS-COCO detection dataset. Finally we finetune the model on the target dataset, which is KITTI in this case.

Such a complex training pipeline comes with a number of hyper-parameters. In our experiments, we aim to keep the settings as simple as possible and ensure that all the models are fairly compared, therefore we use the default settings for both ImageNet  and MS-COCO pretraining. The ImageNet pretraining follows the 90-epoch training scheme in PyTorch examples\footnote{\href{https://github.com/pytorch/examples}{https://github.com/pytorch/examples}}. The MS-COCO detection training follows the open-sourced PyTorch implementation of Faster R-CNN\footnote{\href{https://github.com/ruotianluo/pytorch-faster-rcnn}{https://github.com/ruotianluo/pytorch-faster-rcnn}}. 

When fintuning on  KITTI, the same settings as for MS-COCO are adopted except that we set the number of iterations in proportion to the size of the training set. All models are trained for 25k iterations.

\subsection{Overall results}

Table~\ref{tab:overall} compares mAP, mean Delay and number of operations between a single-model Faster R-CNN detector, a cascaded detector and two CaTDet systems with different configurations. 
The CaTDet systems have higher mAP compared with the single-model detector while requiring 5.15x and 8.67x fewer operations. 
The cascaded systems save slightly more operations than the CaTDet systems but suffer
a loss of 0.5\% to 0.7\% mAP. In our later experiments it is shown that this gap cannot be mitigated even with further increasing the number of proposals/operations.

Here we only consider the arithmetic operations in convolutional layers and fully-connected layers. The tracker and the other layers in DNN models are relatively negligible in terms of either time or operations.

\begin{table*}[t]
\centering
\caption{Comparison on KITTI dataset at Moderate and Hard modes, \textit{Moderate} and \textit{Hard}. (Res10-a, Res50) stands for ResNet-10a as the proposal net and ResNet-50 as the refinement net. mD@0.8 indicates that the delay is tested at an average precision of 0.8. }
\label{tab:overall}
\begin{tabular}{l|lllll}
System        & ops(G)     & mAP(\textit{Moderate}) & mAP(\textit{Hard}) & mD@0.8(\textit{Moderate}) & mD@0.8(\textit{Hard}) \\
                      \hline
Res50, Faster R-CNN      & 254.3 & 0.812    & 0.740  & 2.6   & 3.3  \\
\hline
Res10a, Res50, Cascaded  & 43.2  & 0.807    & 0.733  & 3.2 & 3.8   \\
Res10a, Res50, CaTDet & 49.3  & 0.814    & 0.740  & 2.9  & 3.7    \\
\hline
Res10b, Res50, Cascaded & 23.5 & 0.787 & 0.730 & 4.7 & 5.7    \\
Res10b, Res50, CaTDet & 29.3   & 0.815 & 0.741 & 3.3 & 4.1  \\
\hline

\end{tabular}
\end{table*}

In terms of delay, CaTDet is slightly worse than the single-model detector. CaTDet-A(proposal net: ResNet-10a; refinement net: ResNet-50) incurs an additional delay of 0.3-0.5 frame compared with the single-model Faster R-CNN detector. An interesting observation is that, CaTDet-B(proposal net: ResNet-10b; refinement net: ResNet-50) with a smaller proposal network and fewer operations can still match the original mAP results. Its delay results are, however, much worse. It indicates that the delay statistics is more sensitive to a bad proposal network.

\begin{table*}[t]
\centering
\caption{Operation break-down: The refinement network operates on proposed regions from the tracker and the proposal net. Because of overlaps between these two sources, the two components sum to more than the total number of operations. Unit: Gops. }
\label{tab:breakdown}
\vspace{2pt}
\begin{tabular}{l|l|ll|ll}
\multirow{2}{*}{System} & \multirow{2}{*}{Total} & \multicolumn{2}{l|}{Total break-down} & \multicolumn{2}{l}{Refinement break-down}          \\
 &   & Proposal   &   Refinement    & From tracker & From proposal net \\
 \hline
Res10a, Res50, Cascaded  & 43.2 & 20.7 & 22.5 & / & / \\
Res10a, Res50, CaTDet  & 49.3    & 20.7     & 28.6       & 11.9          & 22.5               \\
Res10b, Res50, Cascaded & 23.5 & 7.5 & 16.0 & / & / \\
Res10b, Res50, CaTDet  & 29.1        & 7.5      & 21.8       & 11.4          & 16.0        \\
\hline
\end{tabular}
\vspace{-5pt}
\end{table*}

\subsection{System analysis}

\textbf{Number of operations break-down.} The number of arithmetic operations for the proposal
and refinement networks are shown in Table~\ref{tab:breakdown}. We also show a further break-down of operations in the refinement network. Due to the overlap of proposals from the tracker and the proposal net, the sum of these components is larger than its actual number of operations.

It is also noticed that with ResNet-10b as proposal net, the refinement network is already dominant. Therefore, further reducing the size of the proposal network, like using ResNet-10c, is not meaningful.

\textbf{Analysis of the proposal network.} We provide an analysis of the proposal network's role in CaTDet.  Table~\ref{tab:single_vs_system} shows different choices of the proposal network with their single-model accuracy and system accuracy. Here ResNet-18 is listed just for comparison purpose. Due to its large size, it is never a serious system design choice.

Considering the great differences of the single-model Faster R-CNN mAP (ranging from 0.542 to 0.687), these four models provide almost identical mAP (0.740-0.742) when acting as the proposal net in CaTDet. On the other hand, a better proposal net makes CaTDet substantially better in the delay metric. CaTDet with ResNet-18 achieves 16\% less delay compared with ResNet-10c. These observations suggest that mAP is not sensitive to the choice of the proposal net, but delay is.

\begin{table}[h!]
\centering
\caption{Comparison of the accuracy of DNN model as (a) a single Faster R-CNN model or (b) the proposal network inside CaTDet system. The refinement net is ResNet-50. mAP and delay are both tested in KITTI's \textit{Hard} mode. }
\label{tab:single_vs_system}
\vspace{2pt}
\begin{tabular}{l|l|lll}
Model   &  Setting   & mAP   & mD@0.8 & ops(G)  \\
                            \hline
\multirow{2}{*}{ResNet-18}  & FR-CNN & 0.687 & 5.9  & 138 \\
                            & CaTDet(P) & 0.742 & 3.5  & 163 \\
                            \hline
\multirow{2}{*}{ResNet-10a} & FR-CNN & 0.606 & 10.9 & 20.7  \\
                            & CaTDet(P) & 0.740  & 3.7  & 49.3  \\
                            \hline
\multirow{2}{*}{ResNet-10b} & FR-CNN & 0.564 & 13.4  & 7.5   \\
                            & CaTDet(P) & 0.741 & 4.0  & 29.3   \\
                            \hline
\multirow{2}{*}{ResNet-10c} & FR-CNN & 0.542 & 15.4 & 4.5   \\
                            & CaTDet(P) & 0.741 & 4.1  & 27.3 \\
                            \hline
\end{tabular}
\end{table}

\textbf{Importance of the refinement network.} The refinement network's accuracy largely determines the overall accuracy of CaTDet. As shown in Table~\ref{tab:refinement}, CaTDet's mAP and delay are very close to the single-model accuracy of the refinement network. It is also noticed that for a less accurate model like ResNet-18, the CaTDet system slightly surpasses the single-model Faster R-CNN, probably due to a better source of proposals.

\begin{table}[h!]
\centering
\caption{Comparison of the accuracy of CNN model as (a) a single Faster R-CNN model (b) the refinement network inside CaTDet system. The proposal net is ResNet-10b. mAP and delay are both evaluated in KITTI's \textit{Hard} mode. }
\label{tab:refinement}
\vspace{2pt}
\begin{tabular}{l|l|lll}
Model   &  Setting   & mAP   & mD@0.8 & ops(G)  \\
                            \hline
\multirow{2}{*}{ResNet-18}  & FR-CNN & 0.687 & 5.9  & 138 \\
                            & CaTDet(R) & 0.696 & 6.0  & 24.4 \\
                            \hline
\multirow{2}{*}{ResNet-50} & FR-CNN & 0.74 & 3.3 & 254  \\
                            & CaTDet(R) & 0.741  & 4.0  & 39.8  \\
                            \hline
\multirow{2}{*}{VGG-16} & FR-CNN & 0.742 & 4.2  & 179   \\
                            & CaTDet(R) & 0.743 & 4.4  & 63.9   \\
                            \hline

\end{tabular}
\end{table}

\textbf{Ablation study of the tracker.} We study how important the tracker is in the CaTDet system. The overall results on KITTI listed in Table~\ref{tab:overall} seem to indicate that eliminating the tracker only incurs minor accuracy loss. However, our experiments show that without the tracker, even if we further increase the number of proposals from proposal net, the accuracy drop cannot be compensated. 

\begin{figure}[h!]
\begin{subfigure}
  \centering
  \includegraphics[width=0.5\textwidth]{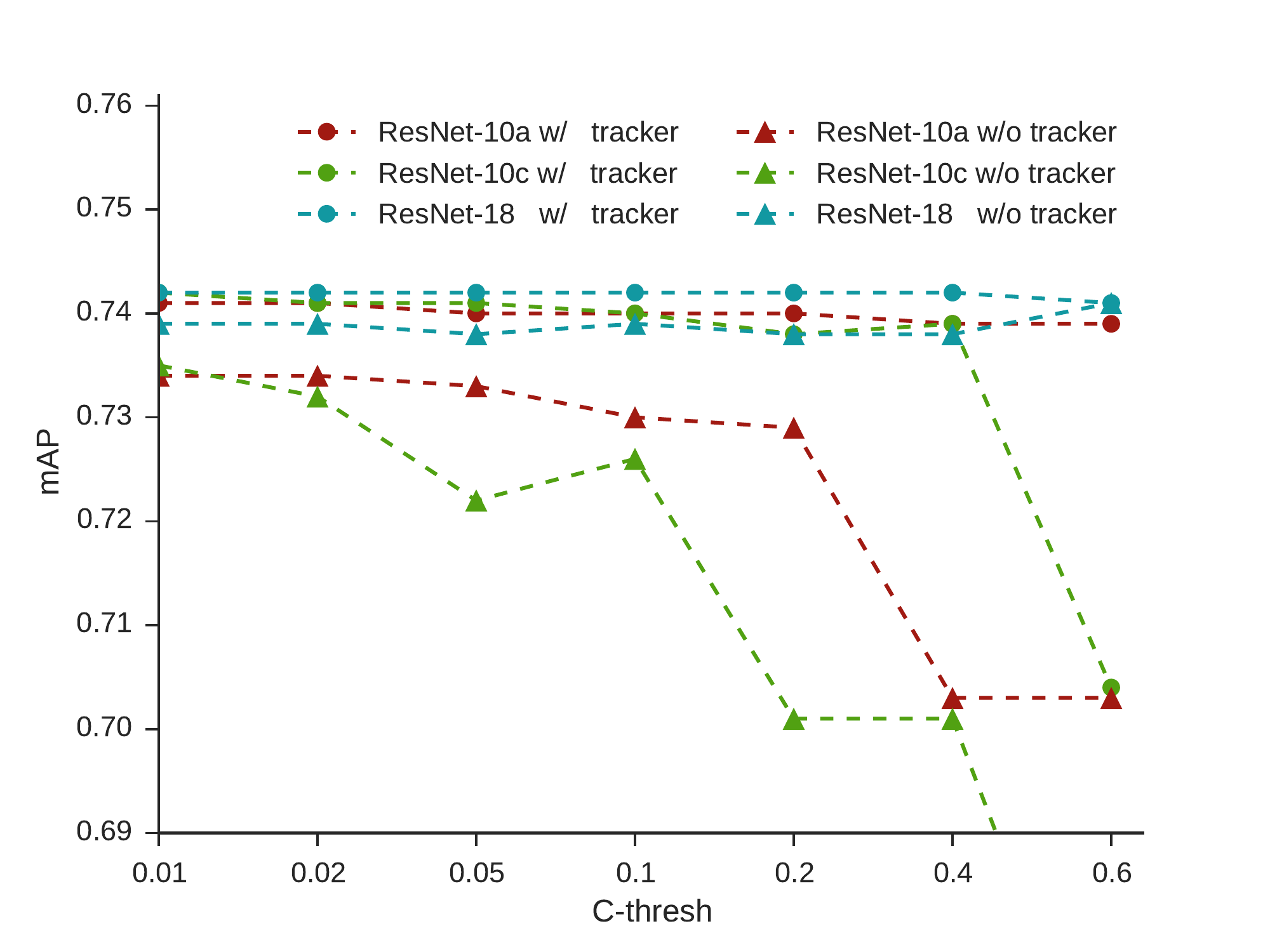}
\end{subfigure}
\begin{subfigure}
  \centering
  \includegraphics[width=0.5\textwidth]{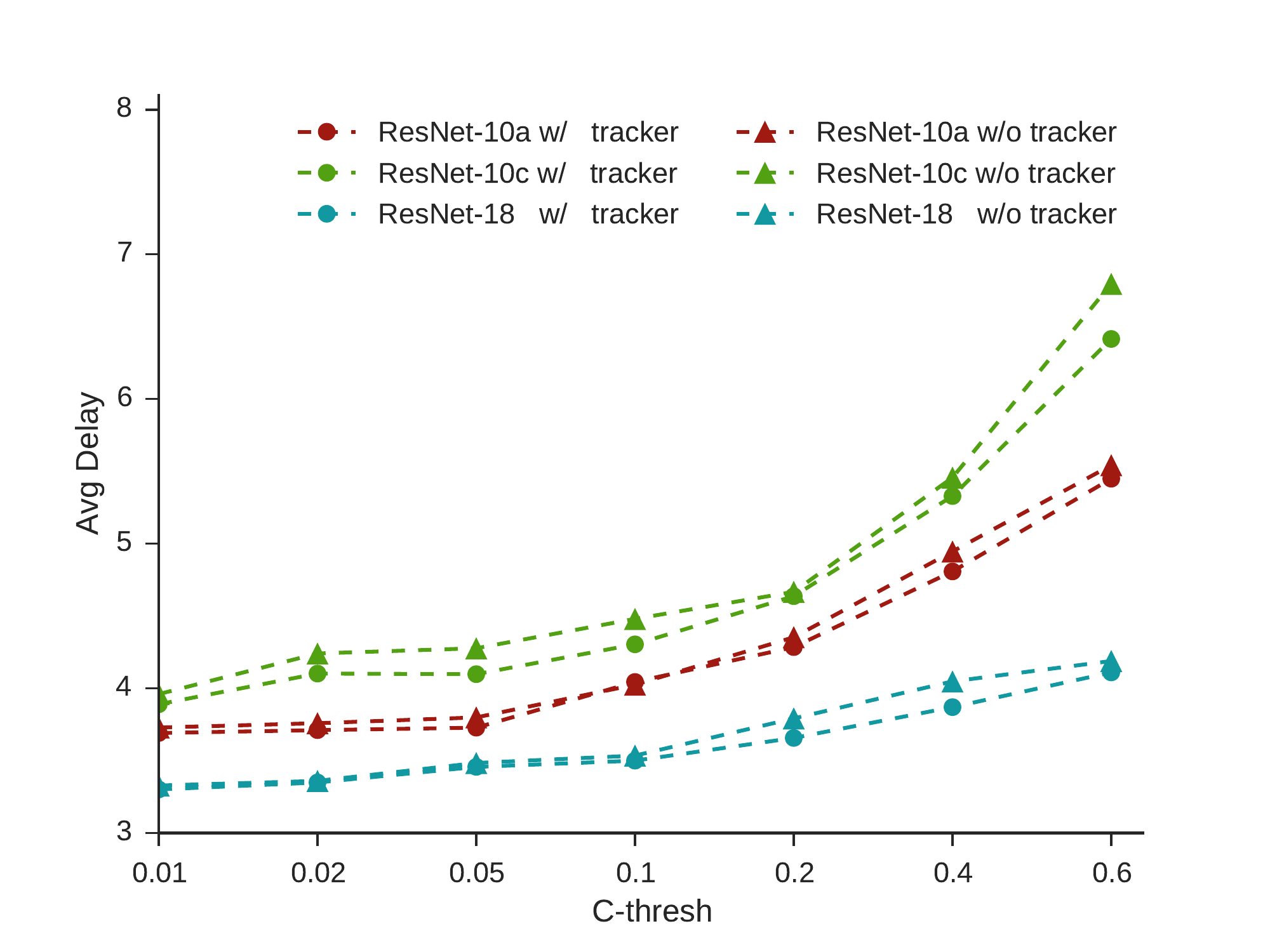}
\end{subfigure}
\caption{Upper: mean Average Precision(mAP) with varying thresholds for the proposal network. Lower: mean Delay at a precision of 0.8(mD@0.8) with varying output thresholds for the proposal network(C-thresh). In the case without a tracker, the system is a typical cascaded system. }
\label{fig:map_tracker}
\label{fig:delay_tracker}
\end{figure}

As shown in Figure~\ref{fig:map_tracker}, in the cases where the tracker exists, varying the C-thresh makes little difference to mAP. 
In the cases without the tracker, none of the cascaded system can match original mAP except Res-18. Also, the cascaded system becomes more sensitive to the choices of the proposal network and the output threshold than CaTDet.
Here C-thresh is the output threshold for proposal net. A higher C-thresh leads to fewer region proposals, reducing the amount of work for the refinement network. 

For the delay metric, both cascaded system and CaTDet are sensitive to the choice of the proposal network and the C-threshold. Figure~\ref{fig:delay_tracker} shows that as C-thresh increases, the average delay gradually increases as well. The main reason is that not enough proposals are fed into the refinement network, causing delayed detection of an object sequence. 

Additional results in Section~\ref{sec:citypersons} further show that removing the tracker greatly harms the performance of the system.

\begin{figure}[!h]
\begin{subfigure}
  \centering
  \includegraphics[width=0.46\textwidth]{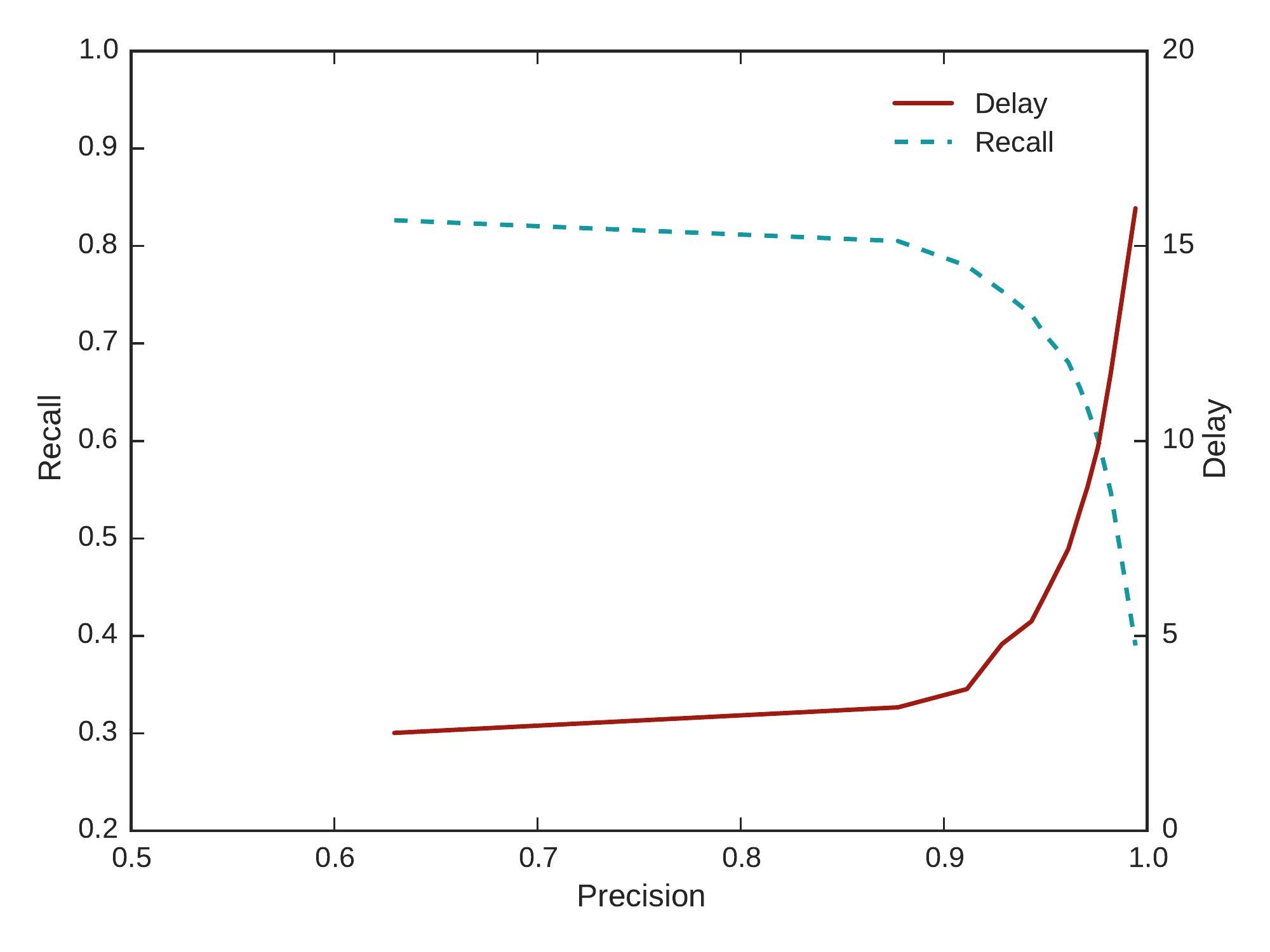}
\end{subfigure}
\begin{subfigure}
  \centering
  \includegraphics[width=0.46\textwidth]{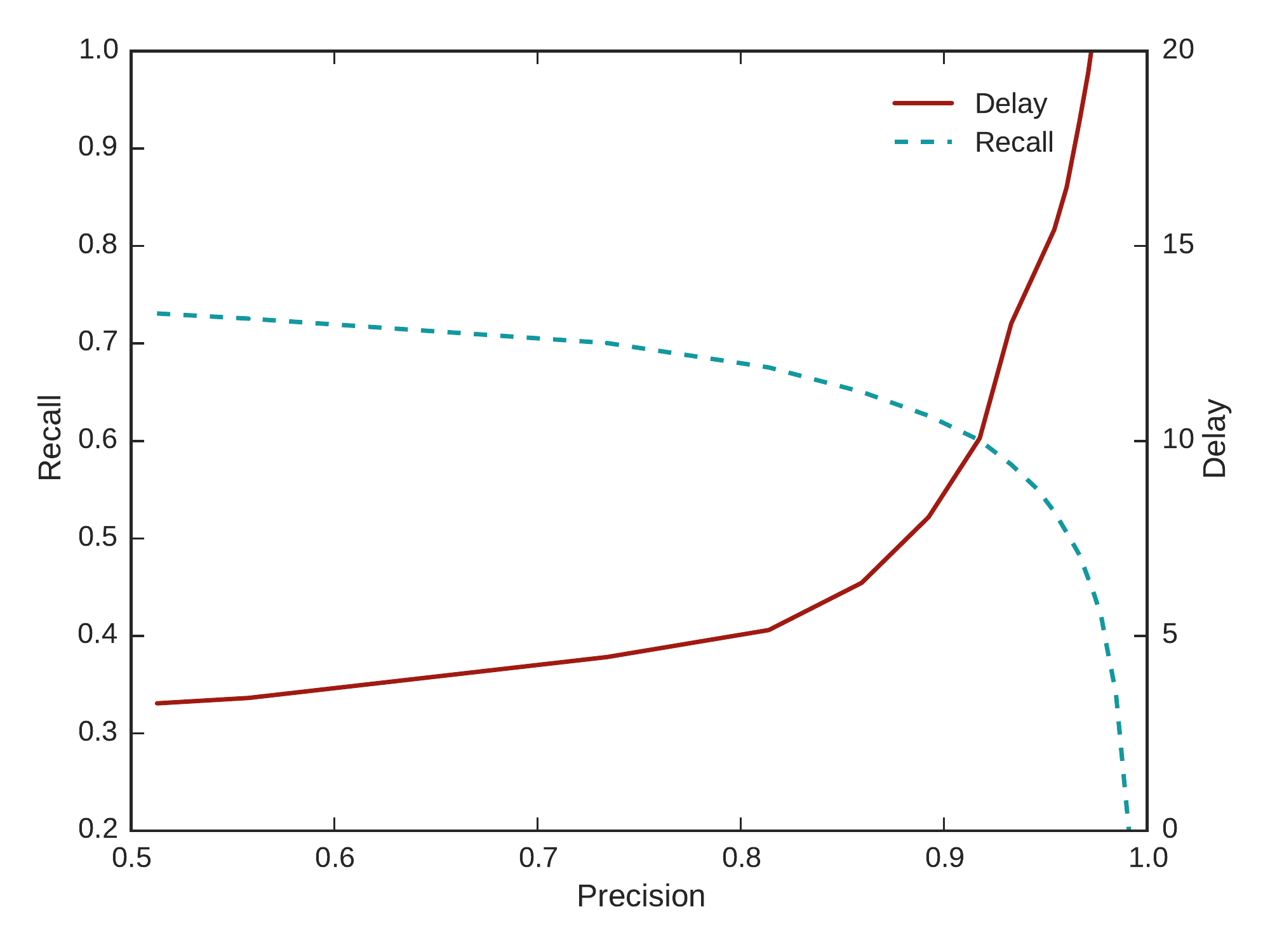}
\end{subfigure}
  \caption{Illustration of how delay, recall correlate with precision. Upper: class \textit{Car}; Lower: class \textit{Pedestrian}. Delay is in unit of frames.}
  \label{fig:ped}
  \label{fig:car}
\end{figure}

\textbf{Visualization of delay-recall correlation}
Figure~\ref{fig:ped} illustrates the trade-offs of recall vs. precision and delay vs. precision. Recall and delay have a strong correlation as the precision changes. Due to fewer number of instances involved in delay evaluation, the delay curve is not as smooth as the recall curve. In this case, pedestrians usually have smaller bounding boxes, which makes them harder to detect.

\section{Experiments on CityPersons}
\label{sec:citypersons}

\subsection{Dataset and training}

CityPersons dataset~\cite{citypersons} provides bounding-boxes level annotations over a subset of CityScapes~\cite{cityscapes}, which is a popular semantic segmentation dataset. As its name suggested, only the class \textit{Person} is annotated in CityPersons. A total of 35016 bounding boxes are labeled in 5000 images.

CityPersons consists of 30-frame sequences at a frame rate of 30 fps and a resolution of 2048x1024. The 20th frame of every sequence is labelled. At training time, we follow the 3-stage pipeline as for KITTI. At inference time, the sparse annotation makes it impossible to evaluate detection delay. Therefore only mAP is evaluated for the CityPersons dataset. The detection system runs on the full sequence, but only the labeled frames are evaluated.

We compute AP for \textit{Person} class following the evaluation protocol of Pascal VOC~\cite{everingham2010pascal}. 
The official CityPersons benchmark follows the protocol of MS-COCO~\cite{mscoco}, which measures mAP under 10 different IoUs(Intersection Over Union) ranging from $0.5$ to $0.95$.

\subsection{Results}

As the CityPersons dataset is sparsely annotated, we only list the mAP in Table~\ref{tab:citypersons}. Due to the fact that frames in CityPersons has much higher resolution than KITTI (2048x1024 vs. 1242x375), the baseline ResNet-50 model has a higher operation count (597Gops vs. 254Gops).

Table~\ref{tab:citypersons} compares the baseline Faster R-CNN model, cascaded system and CaTDet. A significant difference between the results of KITTI and CityPersons is that the cascaded system performs substantially worse on CityPersons. For both configurations shown in the table, eliminating the tracker reduces the mAP by more than 5\%. 

On CityPersons, our proposed system is not able to fully match the original accuracy. CaTDet with a ResNet-10b proposal net and a ResNet-50 refinement net reduces operation count by 13x while degrading mAP by 0.8\%. 

\begin{table}[h!]
\centering
\caption{mAP and number of operations on CityPerson dataset. All the hyper-parameters are kept the same as in KITTI experiments to ensure that CaTDet systems are robust across different scenarios. }
\label{tab:citypersons}
\begin{tabular}{l|ll}
System & mAP & ops(G) \\
\hline
Res50 Faster R-CNN & 0.674 &597 \\
\hline
Res10a, Res50, Cascaded & 0.611 & 79.5 \\
Res10a, Res50, CaTDet & 0.662 & 87.4 \\
\hline
Res10b, Res50, Cascaded & 0.607 & 39.0 \\
Res10b, Res50, CaTDet & 0.666 & 46.0 \\
\hline
\end{tabular}
\end{table}

\section{Conclusion}
In this paper we proposed a new system for detection from video and a new delay metric for delay-critical video applications. On KITTI dataset, the proposed system CaTDet is able to save arithmetic operations of object detection by 5.1-8.7x with no mAP loss and minor delay overhead. On CityPersons dataset, CaTDet achieves 13.0x saving with 0.8\% mAP loss. In addition, our analysis showed that the delay correlates with, but not behaves exactly the same as the recall, therefore requires additional attention for delay-critical detection system design. 

\section{Acknowledgement}
This work is supported by Huawei Technologies Cooperation. We would like to thank Liang Peng and Zhan Xu for comments and discussions that help improve this paper.

\bibliography{draft}
\bibliographystyle{sysml2019}

\section*{Appendix I: Timing Results}
Though we believe that the number of operations is a more general metric to be compared across different platforms, we still list the timing results on the GPU platform for additional information.

Due to the fact that a GPU is very inefficient at processing small batches of data, a simple heuristic was used to merge target regions into a large rectangular area before feeding proposals into the refinement network. Preliminary experiments show that GPU execution time, $T$, of a certain CNN workload, $W$, could be approximated with a simple linear model: $T=\alpha W+b$, where $b$ is estimated to roughly match the execution time of a 400x400 image. A greedy bounding box merging algorithm is thereby derived: two bounding boxes are merged if the merged box has a smaller estimated execution time than the sum of both. In such a way we improves GPU execution efficiency at the cost of increasing the actual workloads.

The experiment is performed on a PC with Maxwell Titan X GPU and Xeon E5-2620 v4 CPU. Both the overall execution time and GPU execution time are measured.  The``Total'' column shows the average process time for a frame, including overheads such as data loading and PyTorch's wrapping. The ``GPU-only'' column shows the GPU kernel time, measured by the NVPROF utility.

\begin{table}[h!]
\centering
\caption{Measured execution time on GPU platform. "Total" stands for the average process time for a frame. "GPU-only" counts the GPU kernel time only.}
\label{tab:gpu_time}
\begin{tabular}{l|ll}
Time(s) & Total & GPU-only \\
\hline
Res50 Faster R-CNN & 0.193 & 0.159\\
Res10a-Res50 CaTDet & 0.094 & 0.042\\
\hline
\end{tabular}
\end{table}

As shown in Table~\ref{tab:gpu_time}, overall execution time is reduced by 2x and GPU time is reduced by 4x. It should also be noted that there is still room for improvement for both overall and GPU time. The CPU time could be hiden by pipelining. GPU execution efficiency could be further improved by either better merging algorithms or other techniques like batching or concatenating small regions.

\section*{Appendix II: Experiments with One-shot Detectors}
CaTDet incorporates Faster R-CNN models for both the proposal network and the refinement network, however, its design principles can also be applied to other detection algorithms, including the popular one-shot detectors like SSD~\cite{ssd}, YOLO~\cite{yolo} and RetinaNet~\cite{lin2018focal}. In this section, we demonstrate CaTDet's efficacy for RetinaNet.

We replace the Faster R-CNN model with RetinaNet as the role of the refinement network. Similar to Faster R-CNN, the workload of RetinaNet can be reduced with selected regions on the original image. RetinaNet only operates at the regions of interest generated by the proposal network and the tracker, thereby reduces the number of operations for both Feature Pyramid Network and Classifier Subnets. 

The training procedure of RetinaNet matches that of Faster R-CNN. We borrowed the code and pretrained model on MS-COCO from the open-sourced repo\footnote{\href{https://github.com/yhenon/pytorch-retinanet}{https://github.com/yhenon/pytorch-retinanet}}. The model is finetuned on KITTI for 5 epochs, following the same settings as in Section~\ref{sec:training}.

\begin{table}[h!]
\centering
\caption{Comparison between single-model RetinaNet and RetinaNet-based CaTDet. Both mAP and delay are tested at Moderate difficulty level. }
\label{tab:retinanet}
\begin{tabular}{l|lll}
Time(s) & ops(G) & mAP & mD@0.8 \\
\hline
Res50-RetinaNet & 96.7 & 0.773 & 6.53 \\
Res10a,Res50-CaTDet & 30.8 & 0.775& 6.33 \\
\hline
\end{tabular}
\end{table}

Results in Table~\ref{tab:retinanet} show that the CaTDet system achieves both better mAP and delay than the single-model RetinaNet detector, meanwhile reduces the number of operations by more than 3x. It suggest that CaTDet could serve as a general framework for CNN-based video detection.

\end{document}